\documentclass[11pt,letterpaper]{article}
\usepackage[pagebackref=true,breaklinks=true,letterpaper=true,colorlinks,bookmarks=false]{hyperref}
\usepackage{graphicx}
\usepackage{algorithm}
\usepackage{algorithmic}
\usepackage{comment}
\usepackage{amsmath,amssymb} 
\usepackage{booktabs} 
\usepackage{color}
\usepackage{url}
\newcommand{\PP}{\mathbb{P}} 
\usepackage{wrapfig}
\usepackage{fullpage}
\begin{document}
\date{}
\title{Contrastive Examples for Addressing the Tyranny of the Majority} 
\author{Viktoriia Sharmanska\\ 
{\small Imperial College London}
\and
Lisa Anne Hendricks\\
{\small DeepMind}\footnote{Work done at UC Berkeley}\\ 
\and
Trevor Darrell\\{\small UC Berkeley}\\ 
\and 
Novi Quadrianto\\ {\small University of Sussex}
}
\maketitle

\begin{abstract}
\it
Computer vision algorithms, e.g. for face recognition, favour groups of individuals that are better represented in the training data. 
This happens because of the generalization that classifiers have to make. 
It is simpler to fit the majority groups as this fit is more important to overall error.
We propose to create a balanced training dataset, consisting of the original dataset plus new data points in which the group memberships are intervened, minorities become majorities and vice versa. 
We show that current generative adversarial networks are a powerful tool for learning these data points, called contrastive examples. 
We experiment with the equalized odds bias measure on tabular data as well as image data (CelebA and Diversity in Faces datasets).
Contrastive examples allow us to expose correlations between group membership and other seemingly neutral features. 
Whenever a causal graph is available, we can put those contrastive examples in the perspective of counterfactuals.

\end{abstract}

\section{Introduction}
\begin{wrapfigure}{rt}{0.5\textwidth}
\begin{minipage}{0.5\columnwidth}
\centering
\vspace{-0.5cm}
\includegraphics[width=0.9\textwidth]{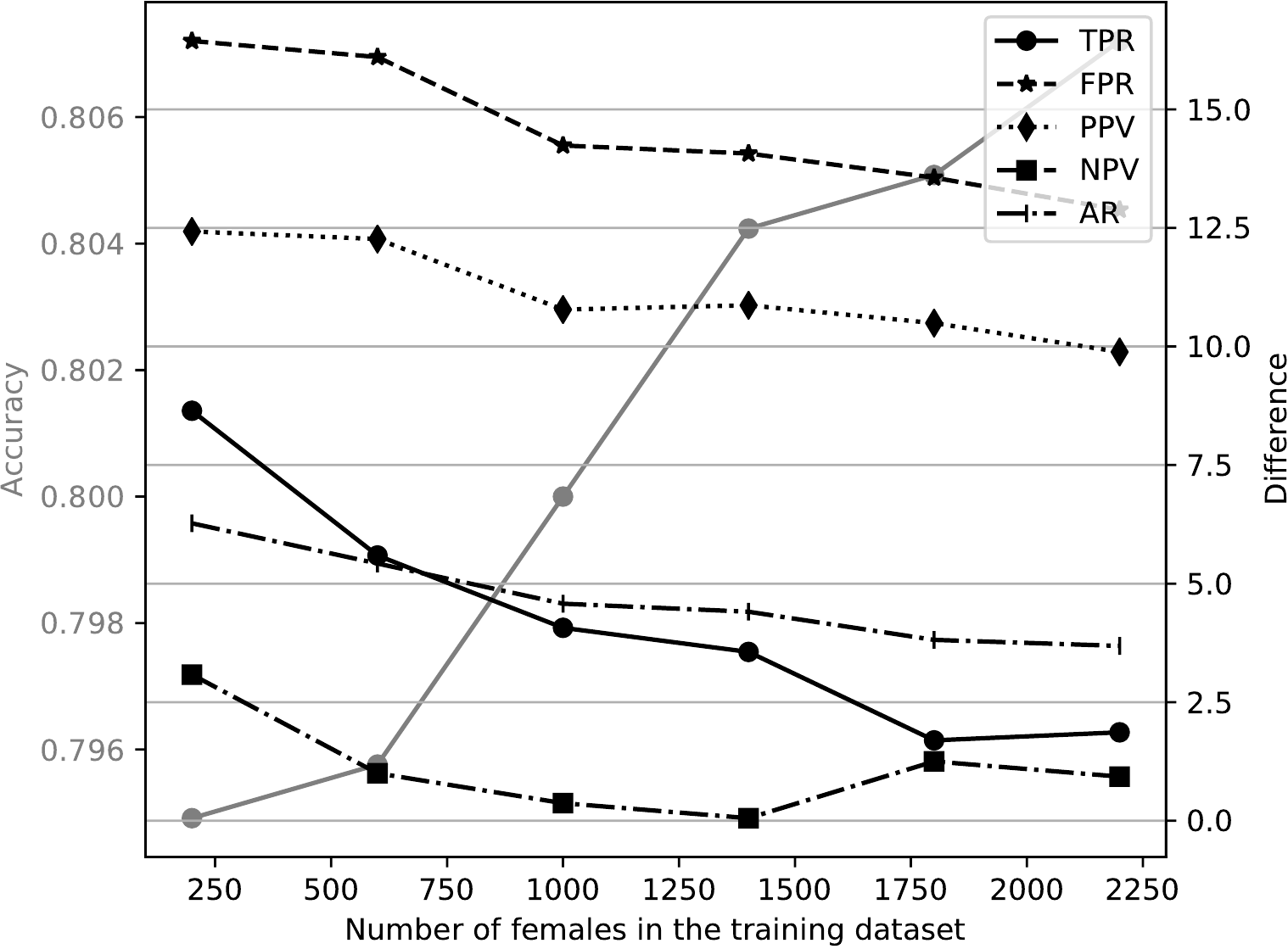}
\caption{An illustration of the tyranny of the majority. As the number of female individuals approaches males ($2200$),  
the gap between males and females in any bias measure decreases (right vertical axis). TPR is true positive rate, FPR is false positive rate, AR is acceptance rate, PPV is positive predicted value, and NPV is negative predicted value.
Left vertical axis is accuracy.}
\vspace{-1.5cm}
\label{fig:illustration}
\end{minipage}
\end{wrapfigure} 

Figure \ref{fig:illustration} illustrates the concept of \emph{the tyranny of the majority} \cite{ChoRot18,HasSriNamLia18} in algorithmic decisions. 
The goal is to predict the income of an individual, without becoming unfairly biased by gender.
The x-axis shows how consistently increasing the number of women in the training set, until the number of men and women is equal, leads to higher accuracy, as well as fairer predictions.
We measure fairness with \textit{several metrics} (difference in true positive rate, false positive rate, acceptance rate, positive predicted value, and negative predicted value between males and females; lower is better for all fairness metrics), and consistently see improved performance as the dataset becomes more balanced.
In all cases the proportion of high/low income individuals remains constant.
Though we can analyze the impact of the tyranny of the majority in a controlled scenario, real data is rarely balanced in such a way.
In contrast, real data reflects a variety of biases including inequality, exclusion, or other forms of inequality.  
The danger is that biases in the training data result in machine learning-based decisions that perpetuate these biases. 
Biased machine learning systems lead to decisions that are allocatively harmful \cite{BarCraShaWal17} to certain individuals because they receive less favourable opportunities (e.g. a loan, an interview) than others.
Using personal data to perpetuate biased decision making is illegal if data refer to personal characteristics that are protected under anti-discrimination legislation, such as, racial origin, gender. 
Furthermore, we continue to expect our AI systems to perform more and more complex tasks, such as facial recognition for law enforcement~\cite{garvie2016perpetual}, or job screening via video interviews (\urlstyle{same}\url{https://www.hirevue.com/}).
Such tasks require building systems which can operate over complex, high dimensional visual inputs while remaining fair and lawful. 

The machine learning community is actively responding to these bias in, bias out problems under the topic of \emph{fairness in machine learning}. 
There are several ways to enforce fairness in machine learning models, \emph{before} the model training phase as a pre-processing step (our method belongs here), \emph{during} the model training phase, or \emph{after} the model training phase as a post-processing step.
Early approaches in fairness models pre-process either the target labels by reweighting \cite{KamCal12}, or the features of the data \cite{ZemeWuSwePitetal13}.
The aim of reweighting is to remove the statistical dependence between the protected attribute and the label. 
Recently, \cite{AgaBeyDudLanetal18} generalize this reweighting method for any definition of fairness that can be formalized via linear inequalities on conditional moments, such as demographic parity (equality of acceptance rates) or equalized odds (equality of true and false positive rates). 
\cite{ZemeWuSwePitetal13} pre-processed the features of the data points by transforming them into a latent space. 
Recent advances in learning fair representations suggest adversarial training \cite{GooPouMirXuetal14} as the means to hide the protected characteristics from the classifier, e.g. \cite{BeuCheZhaChi17,XieDaiDuHovNeu17,ZhaLemMit18,MadCrePitZem18}. 

In this paper, we propose \emph{a simple de-biasing framework} that redefines the reweighting approach based on the concept of intervention or manipulation.
We reweight the original training dataset in which each data point is used twice (for binary protected attributes), however, the second copy is a counterfactual data point by intervening the protected attribute. 
A counterfactual is equivalent to an imaginary intervention on a causal model in which all variables that are not affected by the intervention are assumed to stay at the current observed levels.
Unfortunately, for many applications, we do not have causal models about how the data were generated. 
Instead of counterfactuals that rely on a set of non-testable causal assumptions such as causal graphs and corresponding structural equations, we refer to the data points with intervened protected attributes as contrastive examples. 
To begin with, the contrastive examples can be produced using a statistical matching technique (see for example \cite{RoseRub83}), i.e. for every data point, we find one or more datapoints with similar observable features and the same target label yet having an opposite protected attribute value.
The empirical performance of this matching approach depends on the choice of observable features that act as control variables, and the similarity metric. 
As an alternative, we show that current generative adversarial networks are able to synthesize those contrastive examples. 
Instead of using adversarial learning to \emph{remove} protected attributes from data as done in all previous work, we use them to generate data points with pre-specified protected attributes.
For non-binary protected attributes (e.g. race), each data point will have as many contrastive samples as number of values. 
As an added bonus, contrastive examples can be easily interpreted because they have the semantic meaning of the input (see Figures \ref{fig:adult_attributes} and \ref{fig:faces_attributes}). 
If we want to encourage public conversations regarding fair machine learning systems, interpretability in fairness is an integral ingredient. 
As well, contrastive examples can facilitate rejection learning \cite{Chow57,CorSalMoh16}, allowing models to reject (not make a prediction) if there is no output agreement between original and contrastive examples. 

\paragraph{Related work}
For completeness, we also review methods that enforce fairness during and after training.
To target fairness during training, many methods add fairness terms as constraints to the classifier learning objective function (e.g. \cite{KamAkaAso2012,ZafValRodGum17b, quadrianto2017recycling}). 
The resulting constrained optimisation problem can be turned into a regularised optimisation problem by making use of its Lagrangian dual form, in which the fairness constraints are moved to the objective and act as a regularizer. 
The Lagrangian dual variable in front of the fairness regularizer determines the trade-off between fairness and accuracy. 
Methods for ensuring fairness via a post-processing step, after model training is done, are also available. 
Having access to binary predictions or continuous scores, \cite{HarPriSre16} utilise multiple-threshold rules, where each sub-group has its own threshold, to enforce fairness. 
Though, by having different thresholds for different sub-groups, we suffer from the disparate treatment \cite{BarSel16}, as the classifier makes its predictions based on protected attributes. 
Hence there is also a tension between using \cite{HarPriSre16} and not using \cite{ZafValRodGum17b, quadrianto2017recycling} protected attributes as part of an input to the classifier. 

In our work, we intervene on training samples to build a better dataset. 
Intervention has also been used at test time to interpret/explain AI systems. 
For example, \cite{wachter2017counterfactual} consider generation of counterfactual explanations for classification tasks for tabular input data (data is a vector of various attributes such as gender or race).
After training a model, \cite{wachter2017counterfactual} generate counterfactuals by optimising for a data point which is similar to an input datapoint but yields a different outcome.
Though also considering counterfactual/contrastive examples, our work is fundamentally different because we consider using contrastive examples during training to build better models and our method for generating contrastive examples (nearest neighbour-based and GANs-based) can be applied to more complex input data e.g., natural images. 
Other work in explainable AI aims to discover \textit{causal} relationships~\cite{petsiuk2018rise,kim2017interpretable} in machine learning models through intervention on test samples (e.g., blocking out an object in an image), and observing how the output changes.
We believe that our contrastive examples can also serve as a kind of explanation, both by allowing us to visually observe the difference between original and contrastive examples and by allowing us to understand how outcomes change if a model is provided with a contrastive example.

Recently a few representation learning methods have addressed algorithmic fairness in computer vision \cite{quadrianto2019discovering, SatHofChe18}. 
In \cite{quadrianto2019discovering}, the learned representation is made independent from the protected attribute via Hilbert-Schmidt independence criterion (HSIC). 
In \cite{SatHofChe18}, Sattigeri et al translate the data into a new image-like domain, where the demographics dependency is removed via adversarial network training. 
In computer vision, bias due to gender or ethnicity has been addressed via collecting large diverse datasets \cite{DiF2019, alvi2018turning, wang2019racial} and via learning a domain invariant representation \cite{GanUstAjaGeretal16}. 
Alvi et al \cite{alvi2018turning} define domains as demographic groups and learn image representation that is \emph{confused} about gender and/or ethnicity using domain confusion \cite{tzeng2015simultaneous}. 
Ryu et al \cite{ryu2017inclusivefacenet} explore a two-stage procedure: first learning the demographics representation from an external dataset, and then learning the classifier based on image and demographic representations. 
This approach comes at risk of indirect discrimination, i.e. indirectly conditioning the classification on protected characteristics. 
Wang et al \cite{wang2019racial} learn to transfer knowledge from Caucasian (source) domain to other-race (target) domains via the maximum mean discrepancy (MMD) criterion. 
However, this approach comes with an unethical decision which race to transfer the knowledge from. 
Domain adaptation methods solve distribution mismatch between training and deployment time and do not enforce fairness. 
Our work is also related to prior works in which generative models have been used to create additional data for domain adaptation~\cite{shrivastava2017learning,hoffman2017cycada, tzeng2015simultaneous}, and to model regression tasks such as age progression \cite{zhang2017age}. 
However, our goal is different as we aim to build fair systems and utilize contrastive examples to gain insight into our model's behavior. 

\begin{figure*}[t]
\centering
\begin{minipage}{0.6\columnwidth}
\includegraphics[page=1,width=0.95\textwidth]{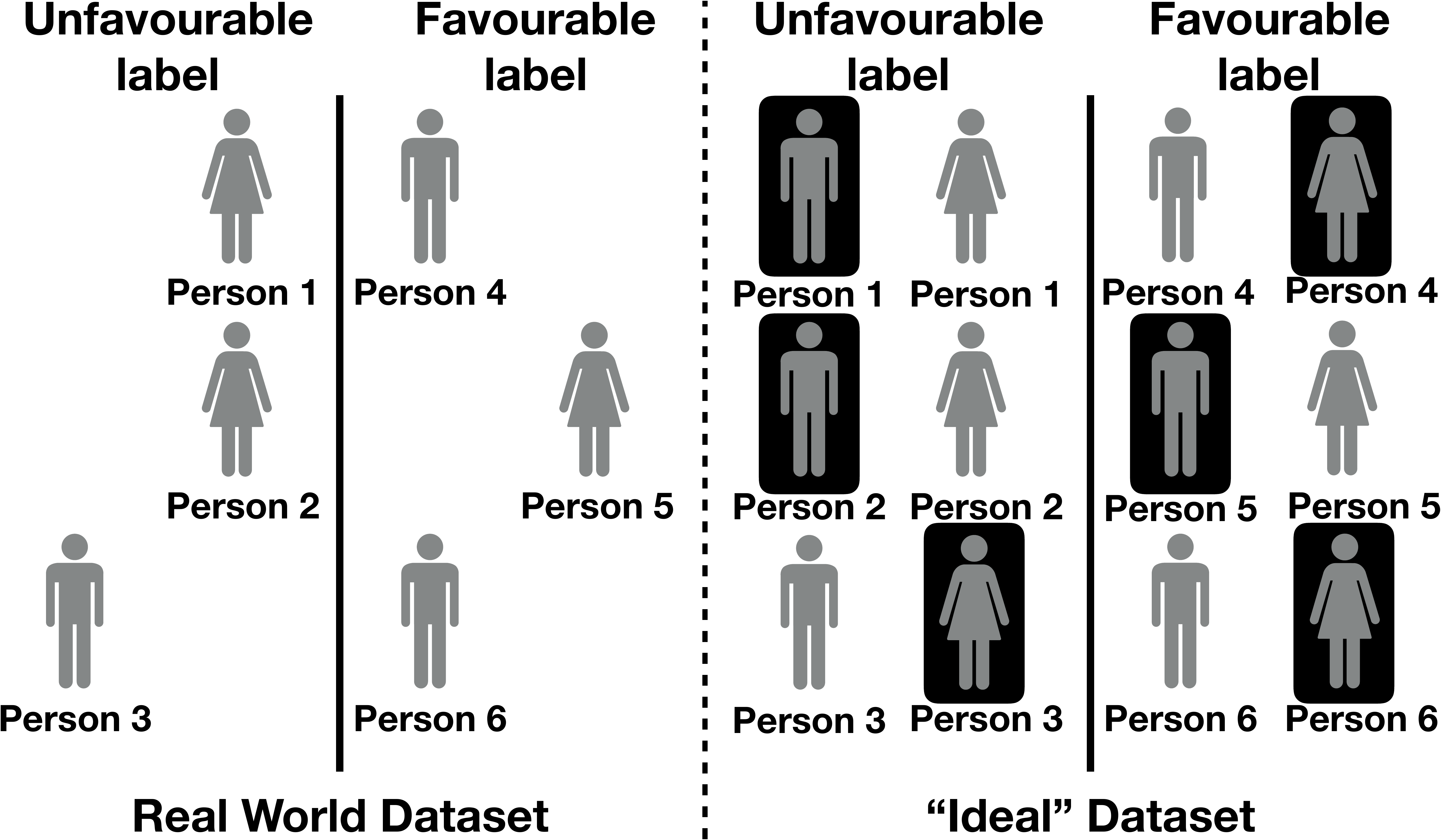}
\end{minipage}
\begin{minipage}{0.3\columnwidth}
\includegraphics[page=2,width=0.95\textwidth]{figures/balanced}
\end{minipage}
\caption{We introduce the ``ideal'' dataset that addresses the tyranny of the majority in a real world dataset (\textbf{left}). 
A favourable label refers to, for example, getting a loan. 
The ideal dataset contains an imaginary data point for each person, i.e. the one inside the black box, whereby we intervene and set the gender attribute to the opposite that is in real life. 
In this paper, we describe two approaches to produce these contrastive data points: 
a generative adversarial network (GAN)-based approach and a nearest neighbour (NN) matching based on a set of features (\textbf{right}). 
While NN contrastive examples are a real person, GAN contrastive examples are not.}
\label{fig:balanced}
\end{figure*}

\section{The ``Ideal'' Dataset}
\label{sec:ideal}
As in the vast majority of fairness in machine learning literature, we focus on the task of batch classification.
An ``ideal'' training dataset is a dataset that for each data point with the corresponding target label, we have an imagined (contrastive) version of it with the protected attribute value, such as gender, is set to the opposite that is in real life while retaining the original target label (Figure \ref{fig:balanced}--\textbf{left}). 
An ideal dataset will produce AI system behaviour that is intuitively fair.
Suppose we have an AI system that is used to produce shortlisted candidates. 
If we had a female applicant and she was invited to the interview, 
had she been a male, she would have also been invited to the interview. 
The same should hold for male applicants. 
Note that an ideal dataset does not affect the distribution of the target label (invited/not invited), but balances group representations within the class (the same amount of females and males got invited/not invited). 
On the other hand, in fairness work, a dataset is called \emph{perfect} if the base acceptance rate is the same for both groups, that is target label is independent of protected attribute. 
If this perfect dataset condition holds, demographic parity and equalized odds can be satisfied simultaneously \cite{KleMulRag17,Chouldechova17}. 
The main hypothesis of this paper is that our ideal dataset with contrastive examples is in fact a perfect dataset, and thus fairness definitions are not mutually exclusive.
The challenge is that we do not have the previously described ideal dataset much in the same way as we can not conduct an ``ideal'' experiment in a study to understand the effect of a particular treatment or exposure \cite{LaaRos11}.
The term ideal dataset is inspired by the concept of an ideal experiment that requires going back in time and intervening on the treatment, but we can also call it a balanced dataset; the two terms can be used interchangeably. 
In this paper, we describe two approaches to produce contrastive data points for generating an ideal dataset: 
a generative adversarial network (GAN)-based approach and a nearest neighbour (NN) matching based on a set of features (Figure \ref{fig:balanced}--\textbf{right}).
We will now detail the GAN-based approach. 

One of the best models to perform image-to-image translation between image domains of different semantic attributes is StarGAN~\cite{StarGAN2018}. 
StarGAN does not require paired images for training and is suitable for learning multi-domain translation. 
Fairness models are typically studied using tabular datasets such as e.g. the Adult Income dataset from the UCI repository \cite{Dua:2017}. 
Next we describe how the StarGAN model can be successfully adapted for synthesis in tabular datasets. 

\noindent \textbf{Data-to-data translation with StarGAN}. 
The StarGAN model consists of two components: a conditional data generator $G$ and a discriminator $D$. 
Given an input sample $x$, and a target sensitive attribute $\bar{s}$, the generator transforms $(x,\bar{s})$ into a synthesized contrastive image $\bar{x}$, which preserves the content of $x$ while changing the target sensitive attribute. 
For example, if gender is a protected attribute, for each \emph{male} sample we create a contrastive example by conditioning the mapping on the \emph{female} attribute as a target characteristic. 
The input to the generator $(x,\bar{s})$ is a depth-wise concatenation of the original $x$ and the spatially replicated target label $\bar{s}$ as an extra channel. 
If the input sample is $224 \times 224 \times 3$ (RGB image), the concatenated input will become $224 \times 224 \times 4$; 
similarly, if the input is $62 \times 1$ (tabular data), the concatenated input to the generator will be $62 \times 2$. 
In case of multiple sensitive attributes, we will add them as extra channels. 
The structure of the generator network follows closely the original StarGAN architecture for image data while replacing 2D convolutions with 1D convolutions for tabular data, and redefining the instance normalization and the output accordingly. 
The output of the generator is a contrastive sample of the same dimension and feature types as the input sample. 
To foster the generator to synthesize fake samples which are both realistic - similar to the ones in the dataset - and reflect sensitive attribute, the discriminator $D$ is trained to perform two tasks: 
i) distinguishing real samples from the synthesized ones and ii) classifying real and contrastive samples w.r.t. their sensitive attribute label. 
The discriminator has the same architecture as in StarGAN except the convolutional layers are replaced with fully connected layers to reflect the classification task in the tabular dataset. 

\noindent \textbf{Data-to-data translation: Component 1.} 
The objective of the conditional GAN-based model for synthesizing realistic contrastive examples can be expressed as a standard adversarial loss:
$\mathcal{L}_{adv} = \mathbb{E}_{x}[log D(x)] + \mathbb{E}_{x, \bar{s}}[log (1-D(G(x, \bar{s}))]$.  
Here the generator $G(x, \bar{s})$ is conditioned on the input sample $x$ and the target sensitive attribute $\bar{s}$ and is trained to minimize the adversarial loss, whereas the discriminator aims to maximize this loss. 
Similarly to StarGAN, for training, we adopt the Wasserstein GAN objective \cite{pmlr-v70-arjovsky17a} with gradient penalty \cite{GulrajaniAADC17}:
$\mathcal{L}_{adv} = \mathbb{E}_{x}[D(x)] - \mathbb{E}_{x, \bar{s}}[D(G(x, \bar{s}))] - \lambda_{GP} \mathbb{E}_{\hat{x}} [(|| \nabla_{\hat{x}} D(\hat{x}) ||_2-1)^2]$,
where $\hat{x} = \alpha x + (1- \alpha) \bar{x}$  is sampled uniformly along a straight line between a pair of a real and a generated samples, $\alpha \sim \mathcal{U}(0, 1)$. 
For tabular data, we might have categorical-valued attributes, however, we note that training a Wasserstein GAN on categorical-valued attributes is done the same way as on continuous-valued attributes \cite{GulrajaniAADC17}.

\noindent \textbf{Data-to-data translation: Component 2.} 
To ensure that the target attribute $\bar{s}$ is present in the synthesized sample $\bar{x}$, an auxiliary classifier $D_{cls}$ is introduced when optimizing both the discriminator and the generator. 
The classifier $D_{cls}$ learns to predict the correct attributes of the real samples by minimizing the attribute classification loss 
$\mathcal{L}^{real}_{cls} = \mathbb{E}_{x, s}[-\log D_{cls}(x,s)]$.
The generator learns to produce contrastive examples $G(x,\bar{s})$ with attributes $\bar{s}$ by minimizing the corresponding classification term on contrastive examples
$\mathcal{L}^{contrastive}_{cls} = \mathbb{E}_{x,\bar{s}}[-\log D_{cls}(G(x,\bar{s}),\bar{s})]$. 

\noindent \textbf{Data-to-data translation: Component 3.} 
%
When learning a translation between the domains, a cycle consistency error is applied, as in \cite{CycleGAN2017,StarGAN2018}, to preserve the content of the input through a cycle  application of $G$, 
$\mathcal{L}_{cyc} = \mathbb{E}_{x, \bar{s}, s}[||G(G(x, \bar{s}), s) - x||_1]$.
%
Here the generator first synthesizes a contrastive example with the target sensitive attribute $\bar{s}$ and then translates it back by conditioning on the attribute $s$ of the original input sample $x$. The $L_1$ norm is used as the reconstruction
loss. 

Finally, the full objective for training the generator and discriminator networks of the StarGAN model for generating contrastive examples can be written as:
$\mathcal{L}_G = \mathcal{L}_{adv} + \lambda_{cls} \mathcal{L}^{contrastive}_{cls} + \lambda_{cyc} \mathcal{L}_{cyc}$,  
$\mathcal{L}_D = - \mathcal{L}_{adv} + \lambda_{cls} \mathcal{L}_{cls}^{real}$,
where $\lambda_{cls}$ and $\lambda_{cyc}$ are the hyper-parameters to control relative importance of different loss terms. 
Note, \emph{both the discriminator and the generator are unaware of the target prediction task} (e.g. predicting the income in the Adult dataset). 

\section{The Model}
For each individual, we have a vector of non-protected features $x\in\mathcal{X}$, a target class label $y\in\mathcal{Y}$, and a vector of special protected features $s\in\mathcal{S}$ (e.g. racial origin or gender).
In almost all current work of fairness, $s$ and $y$ are assumed to be binary.
It is further assumed that one of the class labels (positive label $y = 1$) is desirable, e.g. being accepted for a loan or being invited to an interview. 
\emph{Group fairness} balances a certain condition between groups of people with different protected attributes, $s$ versus $\bar{s}$. 
The term $\hat{y}$ below is the prediction of a machine learning model that, in most works, uses only non-protected features $x$.
Several group fairness criteria have been proposed, a popular one is called equalized odds \cite{HarPriSre16,ZafValRodGum17b}, which demands equality of true $\PP(\hat{y}=1|s,y=1)=\PP(\hat{y}=1|\bar{s},y=1)$ and false positive rates.
We use this measure in our experiments.
 
Given the ``ideal'' training dataset consisting of the original and contrastive examples (Section \ref{sec:ideal}), our proposed de-biasing framework is straightforward. 
It requires only black-box access to a standard classification algorithm (e.g. SVMs, logistic regression, multi-layer neural networks). 
The algorithm does not need to have any knowledge of the desired definition of fairness, and importantly, the algorithm does not use protected attributes as its input.
It simply uses original examples with the corresponding target labels, and contrastive versions of the original data with the original target labels as the training dataset. 
Please refer to Algorithm \ref{alg:contrastive} for the summary of our proposed method.
\begin{algorithm}[t]
\caption{Learning with contrastive examples}\label{alg:contrastive} 
\begin{algorithmic}
\STATE {\bfseries Input} original data $\{(x^n,y^n,s^n)\}_{n=1}^N$, contrastive examples $\{(\bar{x}^n,\bar{s}^n)\}_{n=1}^N$, \\
\STATE Assume access to a classification algorithm that takes training data and can learn an accurate classifier $f:\mathcal{X} \rightarrow \{0,1\}$ (e.g. SVMs, logistic regression)
\STATE Call the algorithm with training data $\{(x^n,y^n)\}_{n=1}^N$ and $\{(\bar{x}^n,y^n)\}_{n=1}^N$
\STATE {\bfseries Return} a classifier $f^{\star}$
\end{algorithmic}
\end{algorithm}
The simplicity of the framework raises both advantages and concerns. 
Which fairness definition are we enforcing?
\cite{KleMulRag17} and \cite{Chouldechova17} showed that most of fairness definitions are mutually exclusive except for special cases when we have a perfect predictor, or when we have a perfect dataset (the base acceptance rate is the same for both groups, i.e. true label is independent of protected attributes).
We do not have perfect predictors, but we can aim to have perfect datasets.
The way we create contrastive examples by intervening on protected attributes, and subsequently assigning the 
original target label to these contrastive examples aims to synthesize this perfect dataset.
If we have a causal graph and we know that the target labels are not affected by the intervention on protected attributes, then the target labels should stay at the current observed levels. 
In general, applying fairness constraints in succession as ``fair pipelines'' do not enforce fairness \cite{DwoIlv18,BowKitNisStraVarVen17}.
Our de-biasing framework relies on contrastive examples, and does not explicitly take a specific fairness definition.
If desired, we could use a fairness-aware classification algorithm as part of the framework. 
%
%
We confirm this experimentally by using the algorithm presented in \cite{AgaBeyDudLanetal18}.

\section{Experiments}
To demonstrate the generality of our de-biasing framework, we apply it to tabular and image data. 

\begin{figure*}[t]
\centering
\begin{tabular}{ccc}
\includegraphics[width=0.3\textwidth]{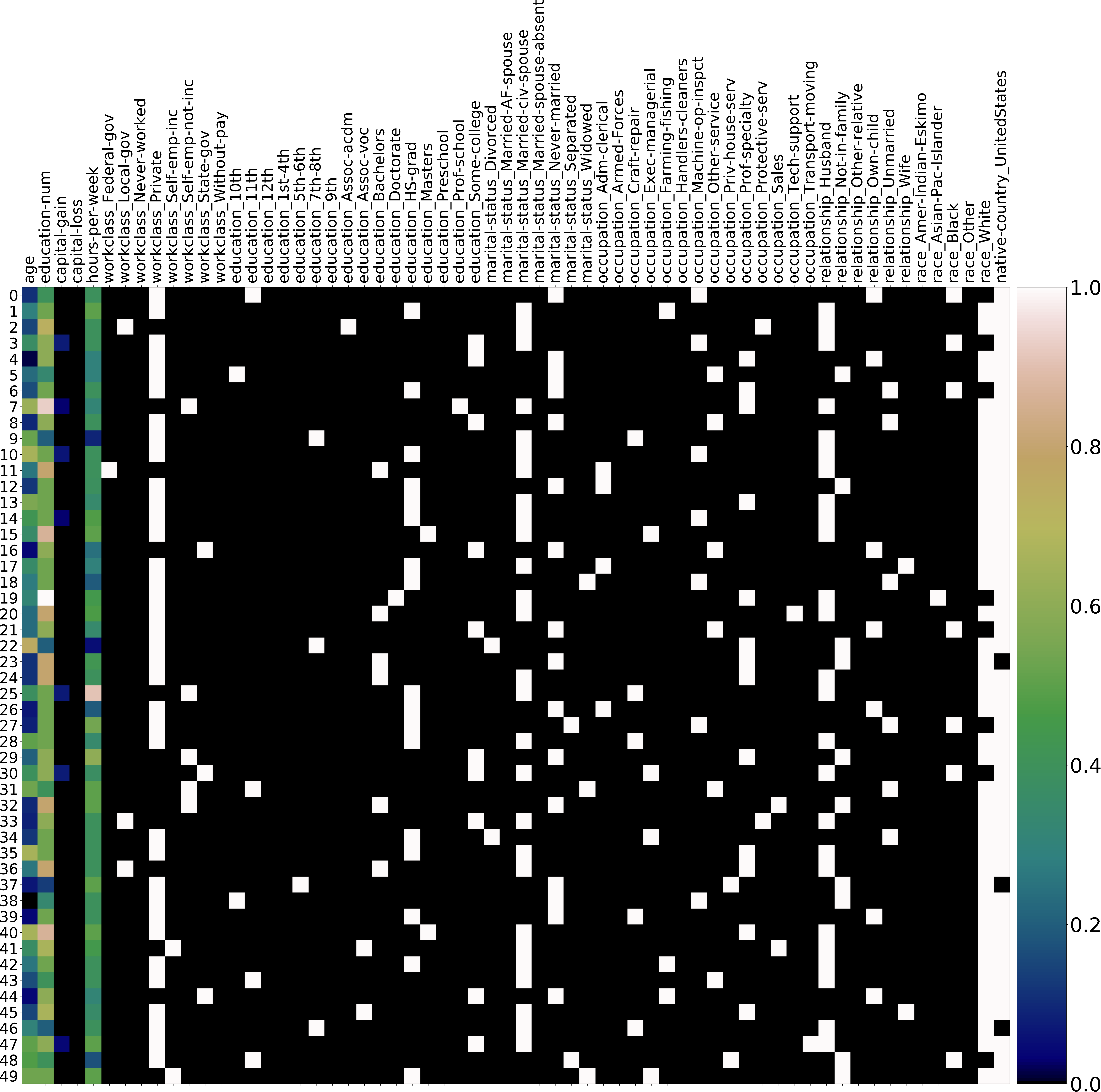} &
\includegraphics[width=0.3\textwidth]{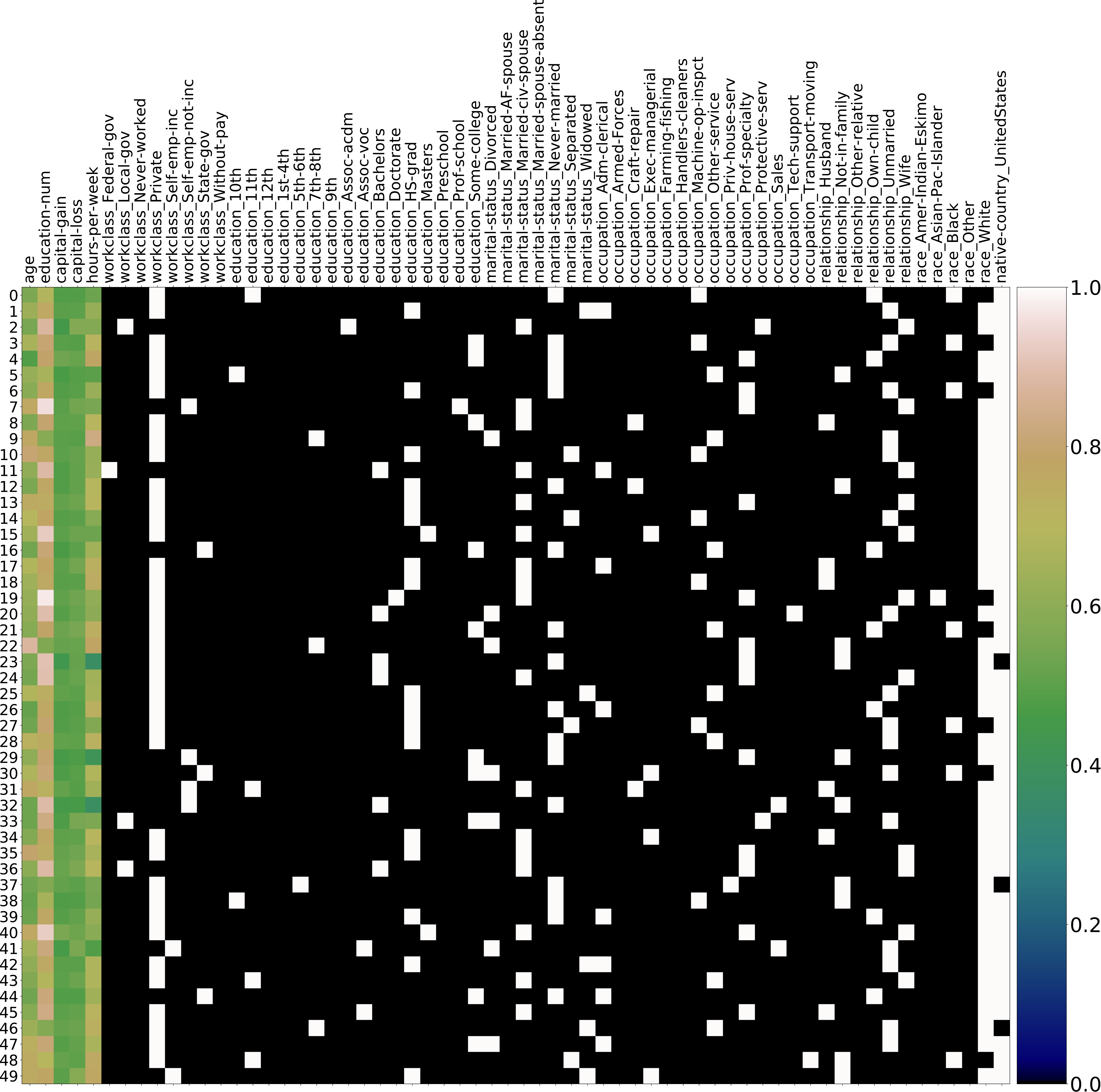}&
\includegraphics[width=0.3\textwidth]{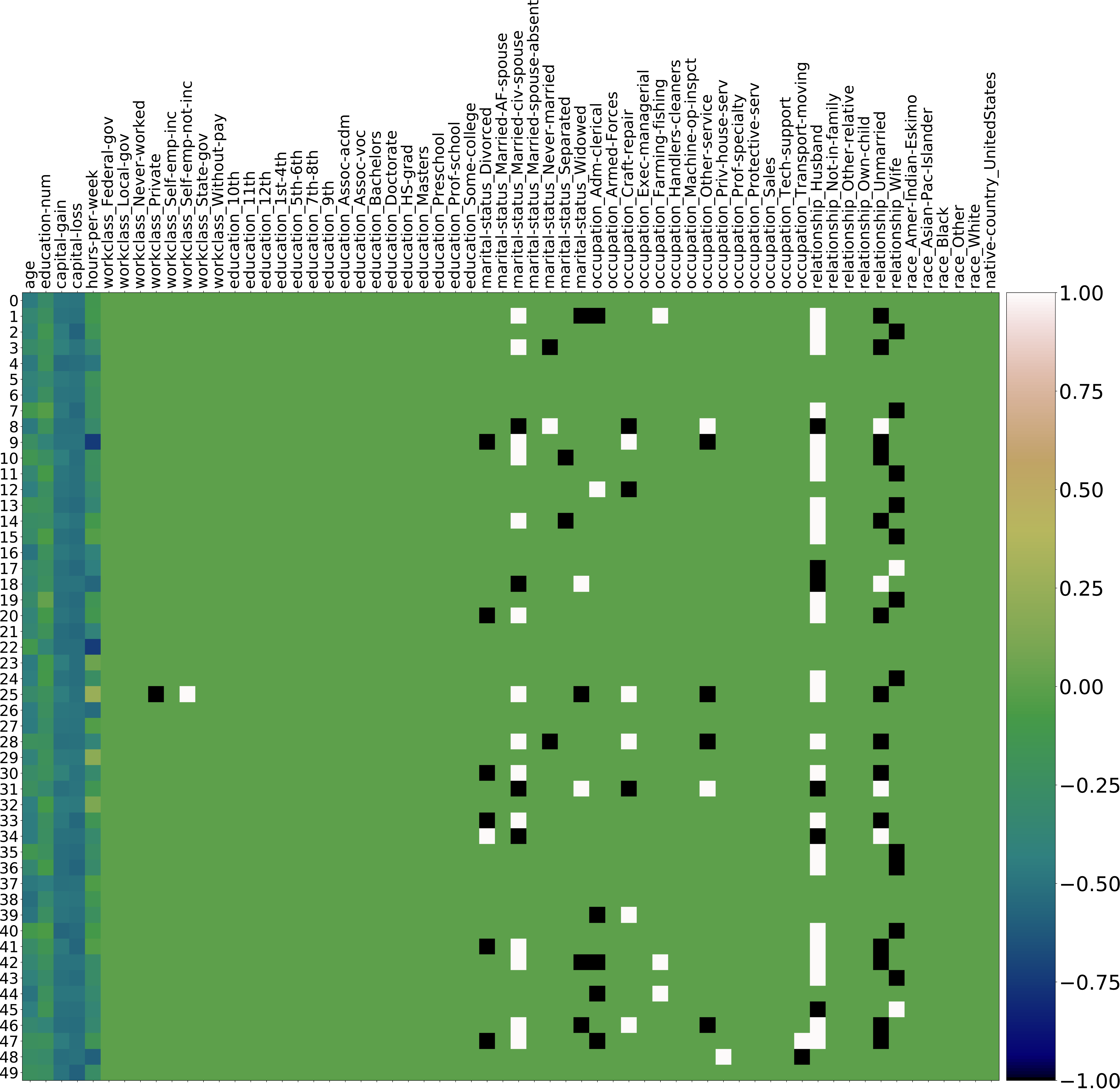}\\
Real examples & Contrastive examples & The difference
\end{tabular}
\caption{Adult dataset: a batch of $50$ real samples (right), their contrastives (center) and the difference between them (left). We stacked the samples for visualization purposes. Each sample has $62$ features (real-valued and binary) along with the label for predicting the income and gender as a protected feature. 
See Section~\ref{sec:visualcontr} for connections between contrastive and counterfactual examples.
We also perform a two-sample test for each feature to understand whether real samples and contrastives 
were drawn from the same distribution or not.}
\label{fig:adult_attributes}
\hspace{0.5cm}
\end{figure*}

\textbf{Tabular data.}  The Adult Income dataset~\cite{Dua:2017} is a common dataset for evaluating ``fair'' machine learning models. 
It consists of 45,222 data instances, of which we use 15,000 instances for test and the rest for training (both the StarGAN model and the de-biasing framework). 
The split procedure was repeated 5 times.  
Each instance is described in terms of $14$ characteristics including gender, education, marital status, number of work hours per week among others, and the goal is to predict whether the income is larger than \$50,000 or not.
%
We transform the representation into $62$ real and binary features along with the sensitive/protected attribute $s$. 
The features can be read from the horizontal axes in Figure~\ref{fig:adult_attributes}. 
Following \cite{ZemeWuSwePitetal13}, we consider gender to be a protected characteristic that is only available during training. 
Our aim is to build models which accurately predict income without unfair bias due to gender.

\textbf{Image data.} 
To explore if our fairness framework transfers to data with high dimensional inputs, we consider the CelebA dataset\footnote{\scriptsize\url{http://mmlab.ie.cuhk.edu.hk/projects/CelebA.html}}~\cite{liu2015celeba} and the Diversity in Faces dataset (DiF)\footnote{\scriptsize\url{https://www.research.ibm.com/artificial-intelligence/trusted-ai/diversity-in-faces/}} \cite{DiF2019}. 
The CelebA dataset consists of over 200K images of celebrities, 20K of which we hold out for testing.
Faces in the CelebA dataset are annotated with a variety of attributes, such as to reflect visual appearance, hair style, accessories, gender, emotional state. 
To simulate vision applications in which visual information is used to predict a favourable attribute (e.g., rating candidates based off a video interview), we predict the \textit{smiling} attribute given an input image.
Where as vast majority of the fairness methods have been demonstrated using a single binary sensitive attribute, our approach can naturally handle multiple sensitive attributes. 
We consider \textit{male/female} and \textit{young/old} as sensitive attributes in this dataset. 
Jointly they form four combinations which will correspond to four sensitive groups when classifying \textit{smiling}.
Our aim is to encourage fair predictions with respect to gender and age. 

The DiF dataset has been introduced very recently and contains nearly a million human face images reflecting diversity in facial features, ethnicity, age and gender. 
The images are annotated with facial landmarks, craniofacial and facial symmetry features, as well as skin color information, age (both continual and discretized into seven age groups), gender. 
We study the effect of the skin color as a sensitive attribute related to ethnicity on the task of age classification that has been shown biased \cite{alvi2018turning} previously. 
Specifically we perform classification into seven age groups: [0-3], [4-12], [13-19], [20-30], [31-45], [46-60], [61+], as provided with the dataset annotation. 
The dataset annotation contains the (estimated) individual typology angle (ITA) that could be used for representing the skin color \cite{del2013variations}. For example, the ITA values below $-30$ correspond to dark skin color, above $55$ to very light skin color. 
We threshold the ITA values at the dataset median ($20$), and use a binarized sensitive feature as skin color attribute. 
We report the results using 500K images for training and the remaining (nearly 412K images) for testing our model on this dataset. 
We discarded a small percentage of original face images due to difficulties in face detection and/or misalignment, or annotation ambiguities.     

\begin{figure*}[!h]
\begin{tabular}{c}
Real images\hfill \null \\
\includegraphics[width=0.1\textwidth]{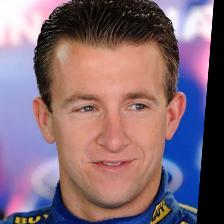}
\includegraphics[width=0.1\textwidth]{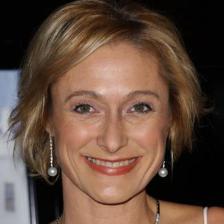}
\includegraphics[width=0.1\textwidth]{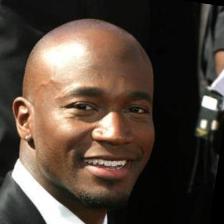}
\includegraphics[width=0.1\textwidth]{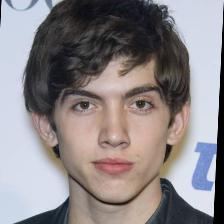}
\includegraphics[width=0.1\textwidth]{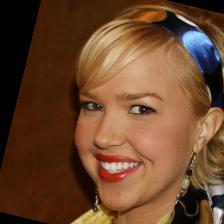}
\includegraphics[width=0.1\textwidth]{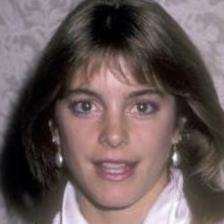}
\includegraphics[width=0.1\textwidth]{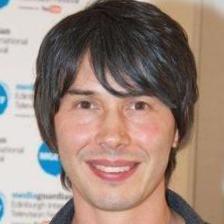}
\includegraphics[width=0.1\textwidth]{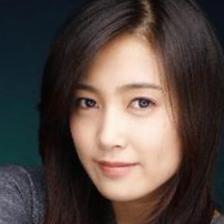}
\includegraphics[width=0.1\textwidth]{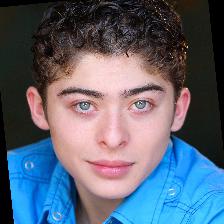}\\
GAN contrastive when intervening on \emph{gender} (\emph{age} unchanged)\hfill \null \\
\includegraphics[width=0.1\textwidth]{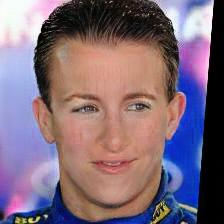}
\includegraphics[width=0.1\textwidth]{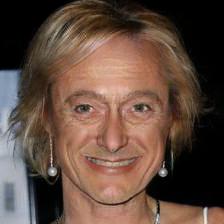}
\includegraphics[width=0.1\textwidth]{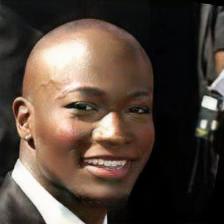}
\includegraphics[width=0.1\textwidth]{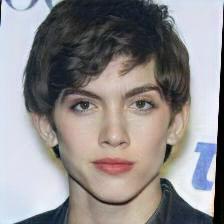}
\includegraphics[width=0.1\textwidth]{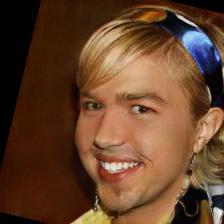}
\includegraphics[width=0.1\textwidth]{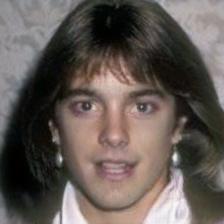}
\includegraphics[width=0.1\textwidth]{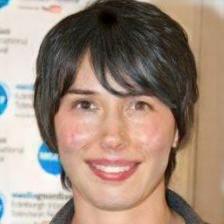}
\includegraphics[width=0.1\textwidth]{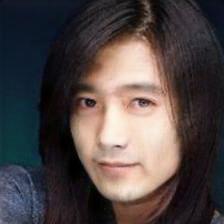}
\includegraphics[width=0.1\textwidth]{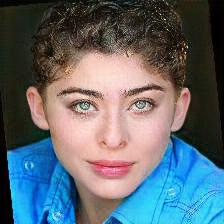}\\
GAN contrastive when intervening on \emph{age} (\emph{gender} unchanged)\hfill \null \\
\includegraphics[width=0.1\textwidth]{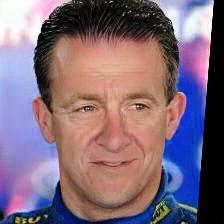}
\includegraphics[width=0.1\textwidth]{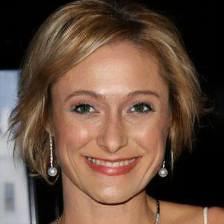}
\includegraphics[width=0.1\textwidth]{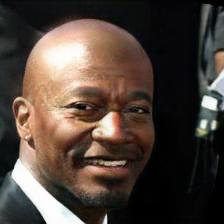}
\includegraphics[width=0.1\textwidth]{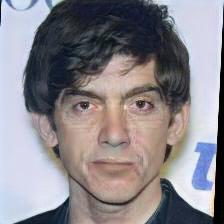}
\includegraphics[width=0.1\textwidth]{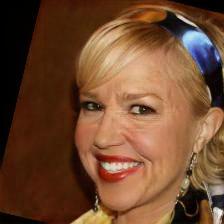}
\includegraphics[width=0.1\textwidth]{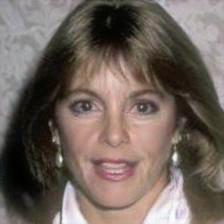}
\includegraphics[width=0.1\textwidth]{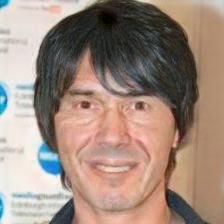}
\includegraphics[width=0.1\textwidth]{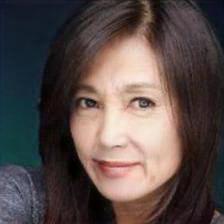}
\includegraphics[width=0.1\textwidth]{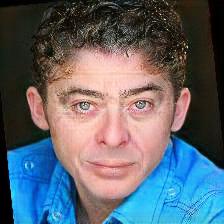}\\
GAN contrastive when intervening on both attributes \emph{gender, age}\hfill \null \\
\includegraphics[width=0.1\textwidth]{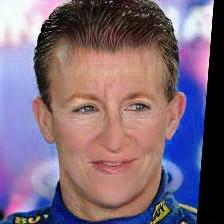}
\includegraphics[width=0.1\textwidth]{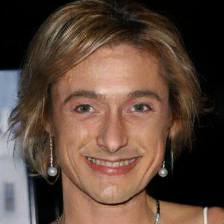}
\includegraphics[width=0.1\textwidth]{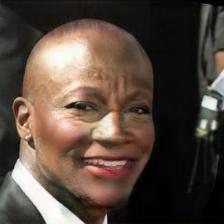}
\includegraphics[width=0.1\textwidth]{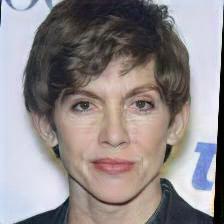}
\includegraphics[width=0.1\textwidth]{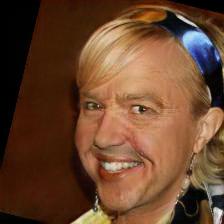}
\includegraphics[width=0.1\textwidth]{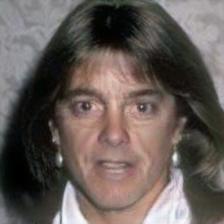}
\includegraphics[width=0.1\textwidth]{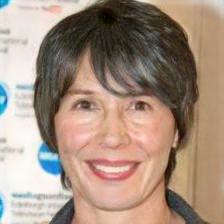}
\includegraphics[width=0.1\textwidth]{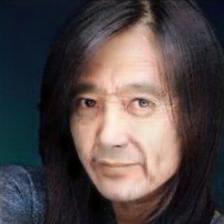}
\includegraphics[width=0.1\textwidth]{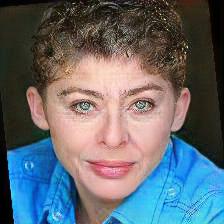}\\\hspace{0.5cm}
\end{tabular}
\caption{Examples of the original images (first row) and their GAN contrastives w.r.t. \emph{gender} and \emph{age} synthesized using the StarGAN model\cite{StarGAN2018}. Both \emph{gender} and \emph{age} are binary attributes. For each real image we synthesize three contrastive examples (second, third and fourth rows). }
\label{fig:faces_attributes}
\end{figure*}
\textbf{Metrics.}
Following prior work in the fairness literature~\cite{friedler2018comparative}, we focus on three important metrics for measuring model performance:  overall accuracy, true positive rate (TPR) parity, and false positive rate (FPR) parity.  
In particular, for fair models, the TPR and FPR should be similar across different groups e.g., the TPR and FPR for predicting age should be the same across light and dark skin color. In case with multiple sensitive attributes, we perform a comparison across all groups, e.g. in CelebA we compare six TPR and FPR pairwise differences across four groups (\emph{young male}, \emph{young female}, \emph{old male}, \emph{old female}). 
By measuring the TPR and FPR differences, we can conclude if our models are fair under the \emph{equality of opportunity criteria} (TPR difference should be small across groups), and \emph{equality of odds criteria} (both TPR and FPR differences should be small across groups).  
We note that it is common for accuracy to decrease as models are more fair according to the equality of odds and equality of opportunity criteria; our goal is to build fair models while minimizing the impact on overall accuracy.

\subsection{Results using tabular dataset}
\label{sec:visualcontr}
We visualize the contrastive examples learned on the Adult Income dataset. 
Figure~\ref{fig:adult_attributes} illustrates a batch of $50$ datapoints stacked for visualization purposes.   
Figure~\ref{fig:adult_attributes} (left) shows the original samples, Figure~\ref{fig:adult_attributes} (middle) shows their contrastive examples, 
Figure~\ref{fig:adult_attributes} (right) shows the difference between them. 
The latter one allows us to analyse which features influence the attribute classifier $D_{cls}$ to predict the opposite gender and drive the generation process of contrastive samples. 
The most prominent is the change in the relationship and marital status, i.e. from \emph{Husband} and \emph{Married-civ-spause} in real samples to \emph{Unmarried} and/or \emph{Wife} in contrastive samples.  
No changes in the workclass or education and only minor changes in occupation have been noticed. 
Similar conclusions hold for the whole dataset. To show this, we perform hypothesis tests using empirical maximum mean discrepancy (MMD) estimates as the test statistic \cite{GreBorRasSch12}, checking whether real and contrastive features were drawn from the same distribution or not.
%
%
Our results confirm that at the significance level $\alpha = 0.01$, we can reject the null hypothesis that the two distributions are the same for all categorical-valued attributes except for workclass, education, race, and nationality. 

\textbf{Connections to counterfactuals.}
\cite[Figure 2b]{NabShp18} postulate a causal graph for the Adult Income dataset.
Based on this causal graph, to remove bias towards females, the direct effect of gender on income, as well as, the effect of gender on income through marital status have to be suppressed.
From Figure~\ref{fig:adult_attributes} (right), and also from the two sample hypothesis testing, we can see that 
contrastive examples have workclass, level of education, race, and nationality variables stay at the observed levels of original data points when we intervene on gender. 
This means when we combine original and contrastive examples to form the ``ideal dataset'' for training an income classification algorithm, the classifier is able to rely on those features (remember that original and contrastive examples have the same income label).
For marital status and family relationship variables, we have the exact opposite. 
From the viewpoint of income classifier, those variables are inconsistent (e.g. married-civ spouse has high income but also widowed given the rest of variables are unchanged), and are therefore should be ignored.
This is encouraging as the causal graph also recommends suppression on the indirect effect of gender mediated by marital status.
\cite{NabShp18}'s causal graph does not include categorical variables race and family relationship.
From our contrastive examples, we can conclude that for gender de-biasing, we should suppress the effect of family relationship but not racial origin. 
We observe some difference in the continuous-valued attributes (capital gain/loss and hours per week) between contrastives and real samples.
Capital gain/loss is not in the causal graph model of \cite{NabShp18}. 
Income level has a causal connection to hours per week, however according to the graph, this connection needs to be suppressed, which has been attempted in our model.  

\begin{table*}[t]
\caption{Results on the Adult Income tabular dataset. The mean and std are computed across $5$ repeats. 
Our approach of enabling the balanced dataset with GAN-based and NN-based contrastive examples is indicated accordingly. Fair baselines are Fair Reduction \cite{AgaBeyDudLanetal18}, and Fair Reweighing \cite{KamCal12}.
}
\label{tab:tabularresults}
\vskip 0.15in
\begin{center}
\scalebox{0.9}{
\begin{tabular}{lccc}
\toprule
Method & Accuracy $\uparrow$ & TPR Diff. $\downarrow$ & FPR Diff. $\downarrow$\\ 
\midrule
\multicolumn{4}{c}{Adult Income dataset}\\
Log. Reg. (real)                 & \textbf{85.16}$\pm$\textbf{0.14} & 7.98 $\pm$1.52  & 7.23$\pm$0.41\\
Fair Reweighing (real)          & 84.37$\pm$0.28 & 14.30$\pm$1.16 &\textbf{1.17}$\pm$\textbf{0.29}\\
Log. Reg. (real and GAN) (ours) & 82.48$\pm$0.44 & {4.95} $\pm${3.67}  & 3.94$\pm$1.33\\
Log. Reg. (real and NN) (ours)  & 85.01$\pm$0.25 & 14.80$\pm$1.90 & 8.20$\pm$0.51\\ 
Fair Reduction (real)       & 83.90$\pm$0.25  & 7.56 $\pm$2.93  & 5.17$\pm$0.36 \\
Fair Reduction (real and GAN)  (ours)    & 82.33$\pm$0.49 & 5.72 $\pm$4.22  & 3.49$\pm$1.60\\
$\ddagger$Log. Reg. (real and GAN & & & \\
    with output consistency) (ours)   & 84.04$\pm$0.42 & \textbf{4.68} $\pm$\textbf{6.37} & 3.33$\pm$1.99 \\
\bottomrule\\
\multicolumn{4}{l}{$\ddagger$classifier only makes a prediction if there is an agreement of outputs between}\\
\multicolumn{4}{l}{original and contrastive examples (at $93\%$ of the test samples).}\\
\multicolumn{4}{l}{}
\end{tabular}}
\end{center}
\vskip -0.5in
\end{table*}
\textbf{Benchmarking on Adult Income dataset.}
%
We use logistic regression as the classification algorithm.
We select the regularization parameter using 10-fold cross validation over 10 possible values $\{500, 100, 50, 10, 5, 1, 0.5, 0.1, 0.05, 0.01\}$.
To ensure that fairness cannot be achieved by simply modifying regularization parameter values,
the two step procedure of \cite{DonOneBenShaetal18} is adopted.
First, we retain the best three parameter values with highest accuracy, then the best parameter is chosen as the one with the lowest bias measure.
We also compare our contrastive method with the latest fairness approach that reduces fair classification to a sequence of cost-sensitive classification problems \cite{AgaBeyDudLanetal18}.
The fairness constraint violation parameter of this reduction approach is searched over 5 values $\{0.005, 0.01, 0.02, 0.05, 0.1\}$. 
Table~\ref{tab:tabularresults} shows that by augmenting the data with contrastive samples, the models consistently become more fair, as shown by lower TPR and FPR rates difference.
This is true for both the logistic regression model and the Fair Reduction model. 

\subsection{Results using image datasets}
On the CelebA dataset, we train the StarGAN model \cite{StarGAN2018} conditioned on \emph{gender} (male/female) and \emph{age} (young/old) binary attributes using 224x224 input images for 60 epochs. 
On the DiF dataset, we extract and align 128x128 face crops from the dataset images and trained the model conditioned on binarized \emph{skin color}(ITA attribute) for 80 epochs.  
Due to the DiF dataset usage restrictions, here we only illustrate contrastive examples on the CelebA dataset -- see Figure~\ref{fig:faces_attributes}. 

\begin{table*}[t]
\caption{
Results on the CelebA HD image dataset. 
We report accuracy of predicting \emph{smiling} using both \emph{gender} (male/female; abbreviated \textit{m} and \textit{f}) and \emph{age} (young/old; abbreviated \textit{y} and \textit{o}) as the two protected attributes $s$. 
Fair baseline is only Fair Reweighing \cite{KamCal12} since Fair Reduction \cite{AgaBeyDudLanetal18} only accepts a single binary protected attribute.
TPR Diff. and FPR Diff. are a mean over six pairwise difference across four groups (using one-vs-one strategy when computing the pairwise difference). 
}
\label{tab:celeba_hd}
\begin{center}
\begin{tabular}{lccc}
\toprule
Method & Accuracy $\uparrow$ & TPR Diff. $\downarrow$ & FPR Diff. $\downarrow$ \\
\midrule
Log. Reg. (real)                  & {89.71}  &6.69 & 6.40   \\
Fair Reweighing (real)            & 89.32  &\textbf{2.74}  & 3.09    \\
Log. Reg. (real and GAN) (ours)   & 88.94  &3.50 & {2.79}   \\
Log. Reg. (real and NN) (ours)    & 88.78  &3.32  & 3.53    \\
$\ddagger$Log. Reg. & & & \\
   (real and GAN with output consistency) (ours)   & \textbf{94.15}  &3.51 & \textbf{2.18}  \\
\bottomrule\\
\multicolumn{4}{l}{$\ddagger$classifier only makes a prediction if there is an agreement of outputs between}\\
\multicolumn{4}{l}{original and contrastive examples (in $17,237$ out of $20,000$ test examples, i.e. $86,185\%$)}
\end{tabular}
\begin{tabular}{cccc|cccc}
\toprule
TPR$_{s=\textit{f},\textit{y}}$ & TPR$_{s=\textit{f},\textit{o}}$ & TPR$_{s=\textit{m},\textit{y}}$ & TPR$_{s=\textit{m},\textit{o}}$ & FPR$_{s=\textit{f},\textit{y}}$ & FPR$_{s=\textit{f},\textit{o}}$ & FPR$_{s=\textit{m},\textit{y}}$ & FPR$_{s=\textit{m},\textit{o}}$ \\
\midrule
\multicolumn{8}{l}{Logistic Regression (real)}\\
\smallskip
94.93& 91.47& 85.04& 83.69 & 20.31& 10.37& 10.14& 7.58\\
\multicolumn{8}{l}{Fair Reweighing (real)}\\
\smallskip
91.36& 89.72& 87.44& 86.64 & 14.35& 8.93& 12.75& 10.45\\
\multicolumn{8}{l}{Logistic Regression (real and GAN contrastive) (ours)}\\
\smallskip
90.61& 90.91& 85.74& 85.54 & 15.01& 10.54& 13.25& 10.33\\
\multicolumn{8}{l}{Logistic Regression (real and NN contrastive) (ours)}\\
\smallskip
91.55& 88.06& 87.60& 85.07 & 12.14& 8.40& 14.80& 10.14\\
\multicolumn{8}{l}{$\ddagger$Logistic Regression (real and GAN contrastive with output consistency) (ours)}\\
\smallskip
96.13& 94.47& 90.56& 90.41 & 8.29& 4.67& 6.38& 4.51\\
\bottomrule\\
\end{tabular}
\end{center}
\vskip -0.5in
\end{table*}
\begin{table*}[!h]
\caption{
Results on the DiF image dataset. 
We report accuracy of predicting seven classes that correspond to age groups: [0-3], [4-12], [13-19], [20-30], [31-45], [46-60], [61+]. We use \emph{skin color} (light/dark; abbreviated \textit{l} and \textit{d}) as the protected attribute $s$. 
Fair baseline is Fair Reweighing \cite{KamCal12}. 
TPR Diff is an average of the per class TPR differences (TPR \emph{light} versus TPR \emph{dark}) across seven classes (age groups). 
FPR Diff is computed similarly. 
}
\label{tab:dif}
\begin{center}
\begin{tabular}{lccc}
\toprule
Method & Accuracy $\uparrow$ & TPR Diff. $\downarrow$ & FPR Diff. $\downarrow$\\
\midrule
Log. Reg. (real)                  & 69.79  &1.55  &0.76\\
Fair Reweighing (real)            & 69.79  &0.98  &0.52\\
Log. Reg. (real and GAN) (ours)   & 69.83  &1.13  &0.64\\
Log. Reg. (real and NN)(ours)     & 69.72  &1.43  &0.63\\
$\ddagger$Log. Reg. & & & \\
   (real and GAN with output consistency) (ours)   &\textbf{72.26}   &\textbf{0.68} &\textbf{0.27}\\
\bottomrule\\
\multicolumn{4}{l}{$\ddagger$classifier only makes a prediction if there is an agreement of outputs}\\
\multicolumn{4}{l}{(in $361 674$ out of $412 556$ test examples, i.e. $87.67\%$). }
\end{tabular}
\scalebox{0.85}{
\begin{tabular}{cc|cc|cc|cc|cc|cc|cc}
\toprule
\multicolumn{2}{c}{[0-3]} &\multicolumn{2}{c}{[4-12]} &\multicolumn{2}{c}{[13-19]} &\multicolumn{2}{c}{[20-30]} &\multicolumn{2}{c}{[31-45]} &\multicolumn{2}{c}{[46-60]} &\multicolumn{2}{c}{[61+]} \\
TPR$_{d}$ & TPR$_{l}$ & TPR$_{d}$ & TPR$_{l}$ & TPR$_{d}$ & TPR$_{l}$ & TPR$_{d}$ & TPR$_{l}$ & TPR$_{d}$ & TPR$_{l}$ & TPR$_{d}$ & TPR$_{l}$ & TPR$_{d}$ & TPR$_{l}$\\
\midrule
\multicolumn{14}{l}{Logistic Regression (real)}\\
\smallskip
87.35& 88.77& 85.83& 85.77& 14.62& 15.28& 69.32& 73.52& 74.87& 72.89& 57.30& 55.71& 54.33& 55.25\\
\multicolumn{14}{l}{Fair Reweighing (real)}\\
\smallskip
87.69& 88.62& 85.55& 85.92& 14.36& 15.56& 70.32& 72.64& 74.26& 73.51& 57.00& 56.02& 54.84& 55.10\\
\multicolumn{14}{l}{Logistic Regression (real and GAN contrastive) (ours)}\\
\smallskip
87.84& 89.11& 86.43& 86.42& 15.79& 16.52& 69.59& 72.76& 73.91& 72.50& 57.66& 56.67& 57.54& 57.84\\
\multicolumn{14}{l}{Logistic Regression (real and NN contrastive) (ours)}\\
\smallskip
83.18& 85.03& 86.44& 87.12& 18.39& 19.69& 69.82& 73.03& 73.88& 72.62& 58.69& 57.40& 53.87& 54.28\\
\multicolumn{14}{l}{$\ddagger$Logistic Regression (real and GAN contrastive with output consistency) (ours)}\\
\smallskip
90.04& 89.93& 89.16& 89.41& 10.02& 9.96& 73.86& 74.36& 76.56& 76.45& 56.69& 57.80& 55.59& 58.18\\
\end{tabular}}
\scalebox{0.85}{
\begin{tabular}{cc|cc|cc|cc|cc|cc|cc}
\toprule
\multicolumn{2}{c}{[0-3]} &\multicolumn{2}{c}{[4-12]} &\multicolumn{2}{c}{[13-19]} &\multicolumn{2}{c}{[20-30]} &\multicolumn{2}{c}{[31-45]} &\multicolumn{2}{c}{[46-60]} &\multicolumn{2}{c}{[61+]} \\
FPR$_{d}$ & FPR$_{l}$ & FPR$_{d}$ & FPR$_{l}$ & FPR$_{d}$ & FPR$_{l}$ & FPR$_{d}$ & FPR$_{l}$ & FPR$_{d}$ & FPR$_{l}$ & FPR$_{d}$ & FPR$_{l}$ & FPR$_{d}$ & FPR$_{l}$\\
\midrule
\multicolumn{14}{l}{Logistic Regression (real)}\\
\smallskip
0.72& 0.97& 2.08& 2.16& 0.65& 0.58& 14.04& 15.69& 19.32& 17.15& 4.37& 3.41& 0.91& 0.76\\
\multicolumn{14}{l}{Fair Reweighing (real)}\\
\smallskip
0.74& 0.96& 2.05& 2.18& 0.63& 0.60& 14.46& 15.30& 18.89& 17.52& 4.32& 3.44& 0.93& 0.75\\
\multicolumn{14}{l}{Logistic Regression (real and GAN contrastive) (ours)}\\
\smallskip
0.73& 0.97& 2.14& 2.22& 0.68& 0.63& 14.08& 15.41& 18.74& 17.07& 4.38& 3.46& 1.08& 0.86\\
\multicolumn{14}{l}{Logistic Regression (real and NN contrastive)(ours)}\\
\smallskip
0.49& 0.69& 2.19& 2.30& 0.91& 0.85& 14.28& 15.48& 18.74& 17.07& 4.58& 3.59& 0.90 &0.70\\
\multicolumn{14}{l}{$\ddagger$Logistic Regression (real and GAN contrastive with output consistency) (ours)}\\
\smallskip
0.70& 0.79& 1.85& 1.90& 0.30& 0.27& 13.84& 14.08& 17.62& 16.69& 3.71& 3.18& 0.71& 0.70\\
\bottomrule\\
\end{tabular}}
\end{center}
\vskip -0.5in
\end{table*}

\textbf{Benchmarking on CelebA dataset. }
We use the ResNet50~\cite{HeZRS16} model pretrained on the large-scale facial dataset \cite{cao2018vggface2} as our feature extractor. 
We use the same evaluation set up as with the tabular datasets and report the results in Table~\ref{tab:celeba_hd}. 
With two sensitive attributes, to rebalance the input dataset, we synthesize three contrastive examples for each input image, i.e. when intervening on \emph{gender}, on \emph{age}, and on both. 
As with the tabular datasets, we observe that training with real and contrastive examples improves fairness (equality of opportunity and equalised odds) with a limited impact on accuracy. Both, TPR and FPR differences get lower when comparing with a logistic regression classifier trained on the original dataset. 
We also observe similar effect with Fair Reweighing. 
The unique selling point for GAN contrastive is that we can actually demand our classifier to only make a prediction if outputs from original test and contrastive test examples agree. 
This is a form of \emph{rejection learning} \cite{Chow57,CorSalMoh16}.
In our setting with two protected binary attributes, this would require the prediction on a test example to agree with its three contrastive counterparts. 
This sounds very strict, will our classifier make enough predictions on the test set?
Actually, yes, it makes predictions on $17,237$ test examples out of $20,000$ ($86,185\%$).
With this output consistency constraint, we can achieve better performance in accuracy and fairness.

\textbf{Benchmarking on DiF dataset. }
We use the same feature extractor as on CelebA and similar evaluation protocol. 
To rebalance the input dataset, we synthesize one contrastive examples for each input image, i.e. when intervening on \emph{skin color}. 
We report the accuracy results of multi-class classification (where seven age groups define seven classes), and equality of opportunity (TPR Diff) and equalized odds (TPR Diff and FPR diff) results of fairness in Table~\ref{tab:dif}. True positive rates difference (TPR Diff) is an average of the per-class TPR differences (TPR \emph{light} versus TPR \emph{dark}) across seven classes. False positive rates difference (FPR Diff) is computed similarly using per-class FPR differences. 
For completeness, we also report the actual TPR and FPR values, as those clearly differ across the age groups. 
As with all other datasets, we observe that training with real and contrastive examples improves fairness. 
TPR differences get lower when comparing with a logistic regression classifier trained on the original dataset. 
In this experiment, training with contrastive examples also slightly improves recognition accuracy. This can be seen from the improvement in TRP values in the groups [4-12], [13-19] and [61+]. 
Fair Reweighing also helps to lower the TPR and FPR differences. 
With the output consistency constraint, we can achieve the best performance in terms of accuracy $72.26\%$ and fairness ($0.68$ and $0.27$ TPR and FPR differences) with an agreement rate $87.67\%$.
Overall, the fairness metric has low discrepancy w.r.t. skin color on this dataset, supporting its main purpose of collecting many diverse face images. 
Nevertheless, the proposed approach can improve over the baselines. 

\paragraph{Visualizing Important Visual Attributes for Classification. } 
\begin{wrapfigure}{tr}{0.5\textwidth}
\centering
\includegraphics[width=0.5\textwidth]{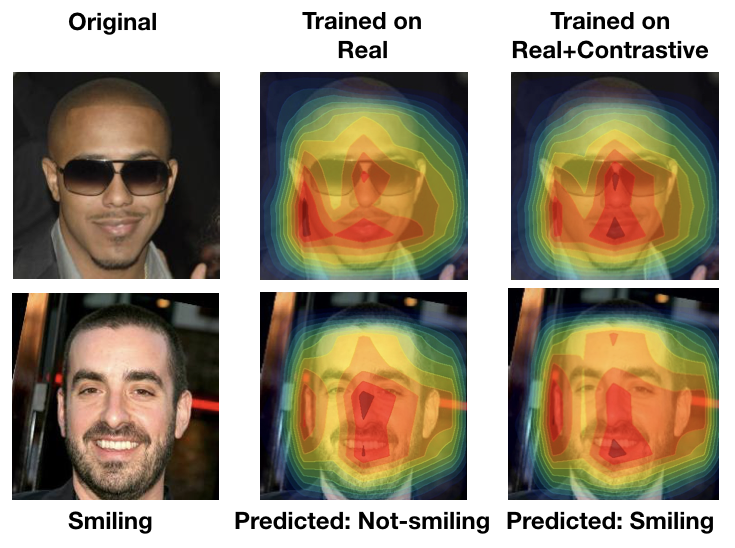}
\vspace{-0.4cm}
\caption{Grad-CAM results of important spatial locations for predicting smiling when trained using real examples and real+contrastive examples.}
\label{fig:gradcam}
\end{wrapfigure}
By producing contrastive examples in the input space, we can qualitatively evaluate what our models focus on during prediction for our test images. 
To visualize which visual attributes are important for our models, we rely on Grad-CAM~\cite{selvaraju2017grad} which traces gradients to determine which locations in an image are most important for a decision.
Figure~\ref{fig:gradcam} shows visualizations for a model trained on real images and a model trained on real and contrastive images, controlling for gender and age as the protected characteristic.
When classifying images, the models tend to focus not only on the mouth area, but also on the center of the face, with important regions including the nose and forehead.
The Grad-CAM visualizations show that the model trained on both real and contrastive images, is more localized on the mouth area of the face, which is  an expected behavior for classifying smiling. 
We noticed such differences were more pronounced in pictures of males rather than females. This is also expected as the male group suffers higher classification error for this task.  
Also we note that the model trained with contrastive images consistently predicts the correct class (\textit{smiling}), suggesting that the model has learned to better ``ignore'' gender and age when making a prediction.

\section{Discussions and Conclusion}
\begin{quotation}
{\flushright\textit{GANs are not just good for causing mischief.}\vspace{-0.25cm}}
\flushright{Karen Hao, MIT Technology Review}
\end{quotation}

In this work we present an intuitive method to train fair AI systems: instead of relying only on the original training data, we intervene on training instances and generate contrastive examples.
By simply building a more ``ideal'' training dataset, trained models become  more fair and less susceptible to uncertainty bias for a variety of input types (tabular data and natural images).
To generate contrastive examples, we build upon progress in image generation.
We demonstrate that current GAN frameworks can easily be extended to very different input modalities (e.g., tabular data), and that contrastive examples provide a means to interpret fair models.
As shown on the Adult Income dataset, our contrastive examples are closely related to counterfactuals.
AI is becoming more and more ubiquitous, and it is imperative that our learning systems can be fair and interpretable.
We believe integrating contrastive reasoning via generative models is an important step towards this goal.
In short, generative models are not just good at causing mischief, but can be a key component in building fair AI systems.

\section*{Acknowledgements}
\noindent VS is supported by the Imperial College Research Fellowship.
NQ is supported by the European Research Council (ERC) funding, grant agreement No 851538. 
We gratefully acknowledge NVIDIA for GPUs donation and Amazon for AWS Cloud Credits.

{\small
\bibliographystyle{ieee_fullname}
\bibliography{bibfile}
}

\end{document}